\documentclass[runningheads]{llncs}
\usepackage[T1]{fontenc}
\usepackage{graphicx}
\usepackage{hyperref}
\usepackage{color}

\usepackage[table]{xcolor}
\usepackage{xcolor}
\usepackage{amsmath}
\usepackage{siunitx}
\usepackage{diagbox}
\usepackage{makecell}
\usepackage{comment}
\usepackage{hyphenat}
\usepackage[caption=false]{subfig}
\usepackage{tikz}
\usetikzlibrary{automata, positioning, arrows, shapes,shapes.geometric}

\definecolor{dark-red}{rgb}{0.4,0.15,0.15}
\definecolor{dark-blue}{rgb}{0.15,0.15,0.8}
\definecolor{medium-blue}{rgb}{0,0,0.5}
\hypersetup{
    colorlinks, linkcolor={dark-red},
    citecolor={dark-blue}, urlcolor={medium-blue}
}

\begin{document}
\title{On the Cost of Evolving Task Specialization \\ in Multi-Robot Systems} 
%
%
\author{Paolo Leopardi\inst{1,2}\orcidID{0009-0005-7064-9344} \and
Heiko Hamann\inst{1,2}\orcidID{0000-0002-2458-8289} \and
Jonas Kuckling\inst{1,2}\orcidID{0000-0003-2391-2275} \and Tanja Katharina Kaiser\inst{3}\orcidID{0000-0002-1700-5508}}
\authorrunning{P. Leopardi et al.}
%
\institute{Centre for the Advanced Study of Collective Behaviour, University of Konstanz, Konstanz, Germany \and
Department of Computer and Information Science, University of Konstanz, Konstanz, Germany\\
\email{\{firstname.lastname\}@uni-konstanz.de} \and
Department of Computer Science and Artificial Intelligence, University of Technology Nuremberg, Germany\\
\email{tanja.kaiser@utn.de}}

\index{Leopardi, Paolo}
\index{Hamann, Heiko}
\index{Kuckling, Jonas}
\index{Kaiser, Tanja Katharina}

\maketitle              
\begin{abstract}
Task specialization can lead to simpler robot behaviors and higher efficiency in multi-robot systems. 
Previous works have shown the emergence of task specialization during evolutionary optimization, focusing on feasibility rather than costs. 
In this study, we take first steps toward a cost-benefit analysis of task specialization in robot swarms using a foraging scenario.  
We evolve artificial neural networks as generalist behaviors for the entire task and as task-specialist behaviors for subtasks within a limited evaluation budget. 
We show that generalist behaviors can be successfully optimized while the evolved task-specialist controllers fail to cooperate efficiently, resulting in worse performance than the generalists. 
Consequently, task specialization does not necessarily improve efficiency when optimization budget is limited.
\end{abstract}
\section{Introduction}

Swarm robotics~\cite{dorigo2021swarm,hamann2018} often relies on simple behaviors (e.g., reactive). 
Task specialization can scale task complexity by dividing labor while keeping individual behaviors simple, as seen in biological systems such as ant colonies~\cite{Wilson_Hoelldobler08,taborsky2025evolution}. 
However, task specialization is only effective when tasks can be partitioned while reducing the requirements for the specialized behaviors.
These behaviors can be hand-coded, learned, or optimized using evolutionary optimization~\cite{bongard13,kuckling2023recent,trianni08}, and task specialization may even emerge naturally~\cite{vanDigg2024emergence,ferrante2015,li2002emergent}. 

In this work, we study the cost of enabling task specialization and whether task-specialized behaviors can evolve under the same evaluation budget.
While most prior work focuses on feasibility, we take first steps toward a cost-benefit analysis of task specialization 
in realistic swarm robotics use cases. 
We argue that task decomposition can destroy synergies in robot-environment interactions, introduce brittle interfaces, and make generalist policies easier to optimize than specialists in some settings.
This mirrors Nolfi's distinction between monolithic and decomposed, modular controllers~\cite{nolfi98,nolfi00}. 
Moreover, for $n$~subtasks, evolving $n$~specialized controllers splits the evaluation budget~$E$ to~$E/n$ per controller, rather than using the full budget for one controller. 

Using the foraging task of leafcutter ants as defined by Ferrante et al.~\cite{ferrante2015}, we compare evolved generalist controllers to task\hyp{}specialist controllers without predefined primitives and under a limited evaluation budget. 
Our results quantify the opportunity cost~\cite{paglieri15}: under the same scarce evaluation budget, generalists achieve comparable or better performance at lower complexity. 
This is particularly relevant for real robots, where sim-to-real transfer often limits evaluation budgets due to computational constraints.

\section{Related Work}

\begin{figure}[t]
    \centering
    \includegraphics[width=0.6\linewidth]{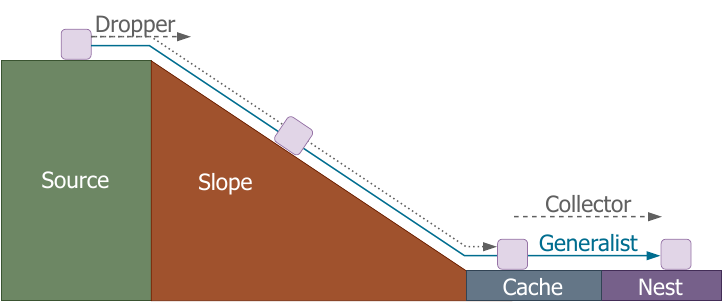}
    \caption{Overview of the foraging task: Robots have to transport objects (squares) from the source across a slope to the nest. The task can either be executed by each robot individually (generalist) or shared between robots (dropper and collector).}
    \label{fig:task_overview}
\end{figure}

Task partitioning and allocation is widely studied in swarm robotics. 
Challenges include coalition formation, resource management, and task interdependencies.
Coalition formation concerns allocating 
groups of homogeneous~\cite {ijspeert2001collaboration} or heterogeneous~\cite{prorok2017impact} robots to tasks, often requiring many tasks to be handled by few robots without causing deadlocks.
Resource management matters when allocation incurs a cost, such as energy~\cite{kernbach2012specialization}, requiring task allocation strategies to balance potential gains against failure risk.
In interdependent tasks, some (sub-)tasks depend on others~\cite{brutschy2014self}, so robots allocated to downstream tasks contribute only after upstream tasks are sufficiently addressed.  
Foraging scenarios, where robots must find and transport resources, include all these challenges.

Inspired by the foraging behavior of insects, threshold models~\cite{krieger2000call} have the robots observe and accumulate stimuli over time.
Once the accumulated stimuli reach a threshold, the robots switch their behavior, e.g., they switch from waiting to foraging~\cite{castello2014foraging,kernbach2012specialization,krieger2000call} or switch between subtasks~\cite{brutschy2014self,lee2020task,pang2019autonomous,pini2011task}.
Pini et al.~\cite{pini2011task} investigated the emergence of task specialization in a threshold-based model.
Robots could either perform a generalist strategy, foraging items from the source into the nest, or learn to partition the task by using a cache.
Depending on the properties of the environment (i.e., the cost of performing the generalist strategy), the swarm learned to either remain generalists or partition the task.

Other works study how task specialization emerges through artificial evolution and reinforcement learning.
Researchers investigate the environmental conditions that benefit specialized behaviors over generalist ones~\cite{ferrante2015,montanier2016behavioral} or how task allocation in homogeneous systems differs from heterogeneous ones~\cite{vanDigg2024emergence,tuci2014evolutionary}. 
Ferrante et al.~\cite{ferrante2015} evolve the control software of a homogeneous robot swarm using predefined low-level behavioral primitives and show that task specialization is evolutionary stable, that is, swarms with successful task specialization do not revert to generalist behavior.
Overall, however, specialized heterogeneity outperforms homogeneous systems~\cite{montanier2016behavioral,tuci2014evolutionary}, with heterogeneity either being predefined~\cite{vanDigg2024emergence} or emerging during the design process~\cite{montanier2016behavioral}. 
Hiraga et al.~\cite{hiraga2018evolving} evolve controllers for manually defined subtasks and an arbitrator controller for subtask selection.
This hierarchical structure outperforms controllers based on a single artificial neural network (ANN) by favoring collaborations among robots.
Furthermore, it overcomes the bootstrapping and deception problem shown by the single ANN-based controller. 

\section{Task, Experimental Setup, and Methods}

\subsection{Task Description}
\label{sec:task_description}

We study a foraging task inspired by leafcutter ants harvesting leaves from trees~\cite{ferrante2015}. 
Our environment has four areas (Fig.~\ref{fig:task_overview}): source, slope, cache, and nest. 
\(M\)~objects~\(O_m, \, m \in M\) are initially randomly distributed in the source. 
A~swarm of \(N\)~robots has to retrieve objects from the source and transport them to the nest. 
When dropped on the slope, objects slide down into the cache. 
Robots can exploit the slope to deliver objects faster to the cache. 
The performance of the swarm is measured by the number of objects~\(C^T_\text{Nest}\) brought to the nest within an evaluation time~\(T\).
Following Balch~\cite{balch2002taxonomies}, we formalize \(C^T_\text{Nest}\) by 
\begin{align} \label{equ:c_T}
    C^T_\text{Nest} &= \sum_{m=1}^M H(O_m, t_0 + T, \text{Nest}), \\
    H(O_m, t, \mathcal{A}) &= 
    \begin{cases}
    1,& \text{if object \(O_m\) is in area \(\mathcal{A}\) at time~$t$} \\
    0,              & \text{otherwise} 
\end{cases}
\end{align}
with start time~\(t_0\) and  \(H(O_m, t, \mathcal{A})\)
indicating if object~\(O_m\) is in area~\(\mathcal{A}\) at time~\(t\).

The foraging task can be performed by two strategies:
(i)~each robot executes the full task alone, that is, it collects an object from the source and brings it back to the nest (\textit{generalist}), or (ii)~robots split the task into two interdependent subtasks.
\textit{Droppers} specialize in moving objects to the cache by exploiting the slope, and \textit{collectors} specialize in transporting them from the cache to the nest.

\subsection{Experimental Setup}

We run our experiments in the open-source simulator Gazebo~\cite{koenig2004design}.

\paragraph{Arena. }

We use an \(4\,\text{m} \times 7.5\,\text{m}\)~arena with a \(1.5\,\text{m}\)-long nest, \(1\,\text{m}\)-long cache and source areas, and a \(4\,\text{m}\)-long slope. 
Three light sources indicate the nest.
Cylinders (height: \SI{0.06}{m}, radius:~\SI{0.1}{m}, weight:~\(1\,\text{kg}\)) are used as objects.
Seven objects are randomly placed in the source at the beginning of each experiment.

\paragraph{Robot. } 

We use the differential-drive TurtleBot~4 and extend it with an emulated grasping mechanism, where objects automatically attach and release in specific areas depending on the behavior. 
Generalists and droppers attach objects in the source and release them in the nest and on the slope, respectively, while the collector attaches objects in the cache and releases them in the nest.
Each robot can grasp only one object at a time, however, multiple objects can be pushed at once.
We use the robot's seven infrared (IR) sensors normalized to~\([0,1]\), and its \(360^\circ\) 2D~LiDAR capped to~\SI{1}{m}.   
To reduce the number of sensor values, we divide the LiDAR readings into eight sectors and take the minimum reading of each sector. 
To detect the area a robot is currently in, we emulate four ground sensors. 
Each outputs a discrete value based on the current area: \(0.2\)~for the nest, \(0.4\)~for the cache, \(0.6\)~for the slope, and \(0.8\)~for the source.
In addition, we place three light sensors on the top of the robot to detect the current light gradient relative to it.
The possible light directions---front, right, left, and back---are mapped to the values \(0.2\), \(0.4\), \(0.6\), and \(0.8\), respectively.
We add Gaussian noise to the IR sensors and the LiDAR. 
For the ground sensors, the correct area encoding is returned only with an \SI{85}{\%} probability.
The likelihood of returning the encoding of another area decreases the farther the area is from the true area.

\paragraph{Control Architecture. }

Each robot is controlled by a fully-connected feedforward ANN,  
whose weights are optimized through evolution.
It has one hidden layer of eight neurons, outputs a linear velocity~$v$ and an angular velocity~$\omega$, and has 21 inputs~$s_1, \dots, s_{21}$: $s_1, \dots s_7$ are the normalized IR readings, $s_{8}, \dots, s_{15}$ are the preprocessed LiDAR values, $s_{16}, \dots, s_{19}$ are the ground sensor readings, $s_{20}$ is the preprocessed light sensor value, and $s_{21}$ is a binary value indicating whether the robot is currently grasping an object.

\subsection{Evolution} \label{sec:evolution-setup}

We optimize ANN weights using a simple evolutionary algorithm~\cite{eiben2015introduction}, where genomes directly encode for weights.
Initial weights are uniformly sampled from $[-0.5, 0.5]$.
We use a population size of~\(100\), tournament selection with a tournament size of \(2\), elitism of \(1\), one repetition, no crossover, and run evolution for \(100\)~generations.
Each gene is mutated with a \SI{2}{\%} probability using Gaussian mutation with zero mean and standard deviation of \(0.2\). 
Initial robot poses and object placement are randomized across generations. 
As the foraging task does not require close robot collaboration, we argue that the three robot behaviors (see Sec.~\ref {sec:task_description}) can be evolved individually. 
Thus, we use only a single robot during optimization. 
We vary the simulated evaluation time~\(T_\text{eval}\) and the configuration of the environment (i.e., initial robot pose and initial object placement) based on the behavior being evolved. 
For the generalist, we set \(T_\text{eval}=4\,\text{min}\), the robot is initialized in either the nest or the cache, and objects are positioned in the source.
For the dropper and collector, we set \(T_\text{eval}=1\,\text{min}\).
The dropper starts in the source with objects, while the collector starts in either the nest or the cache, with objects placed in the cache.
We use longer evaluations for the generalist because they need to traverse the full arena to deliver items to the nest. 

For the generalist and the collector behaviors, fitness is defined as the number of objects~\(C^T_\text{Nest}\) in the nest at the end of the run at time~\(T_\text{eval}\) (see Eq.~\ref{equ:c_T}). 
We define fitness~\(F_G\) for the generalist and fitness~\(F_C\) for the collector behaviors as
\begin{equation} 
    F_G = F_C = C^T_\text{Nest} = \sum_{m=1}^{M} H(O_m, t_0 + T_\text{eval}, \text{Nest})\;.
    \label{eq:F_G}
\end{equation}
For droppers, we use the number of objects on the cache, as defined by 
\begin{equation}
    F_D = \sum_{m=1}^{M} H(O_m, t_0 + T_\text{eval}, \text{Cache})\;.
    \label{eq:F_D}
\end{equation}
These aggregate fitness functions measure task success without rewarding specific behaviors~\cite{nelson2009fitness}. 
We ran evolutions on two PCs, one with an AMD Ryzen~9 7900 CPU and a NVIDIA GeForce RTX 4070/4080, and another one with an Intel~i5 and an NVIDIA GeForce RTX 4070 Ti Super.
Each run took about 10 days for a generalist and 2 days for each specialist.
We limited experiments to three independent runs each for the generalist, dropper, and collector.

\subsection{Post-Evaluation} \label{sec:post-evaluation}

We post-evaluate both foraging strategies (see Sec.~\ref{sec:task_description}) using the best genomes from the last generation of each evolutionary run. 
Controllers are tested with \(N=2\)~robots in two settings:
(i)~homogeneous pairs of generalists sharing the same evolved weights, and (ii)~mixed pairs composed of one dropper and one collector.
We combine the droppers and collectors in the order of their evolution (i.e., the first evolved dropper with the first evolved collector, etc.). 
In addition, we evaluate the best of the three evolved droppers and collectors together.
We test each configuration in ten random trials with \(7\)~objects and an evaluation time of~\(5\,\text{min}\). 
In each trial, one robot begins in a random pose in the nest (generalist or collector) while the other starts in the source (generalist or dropper).

\section{Results}

\subsection{Evolution}
\label{subsec:evolution_results}

\begin{figure}[tb]
\centering
\subfloat[generalist \label{fig:fit_generalist}]{\includegraphics[width=0.33\textwidth]{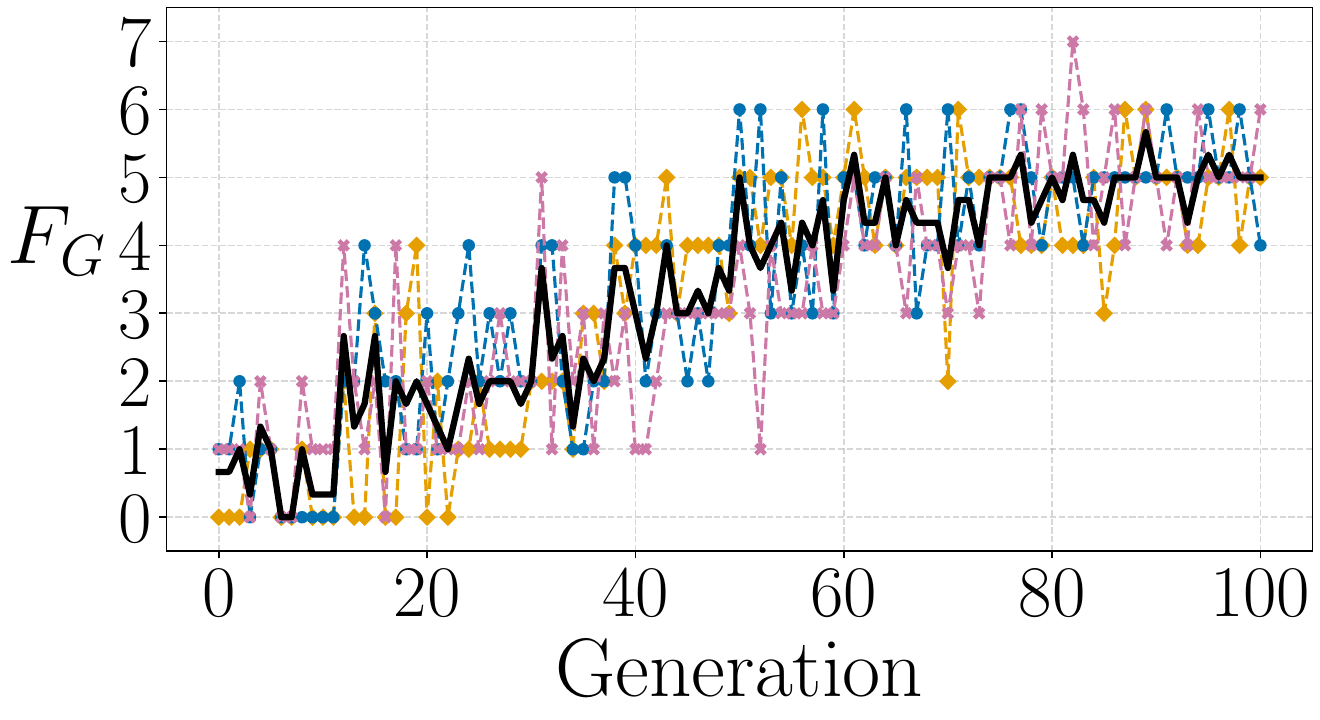}} \hfill
\subfloat[dropper \label{fig:fit_dropper}]{\includegraphics[width=0.33\textwidth]{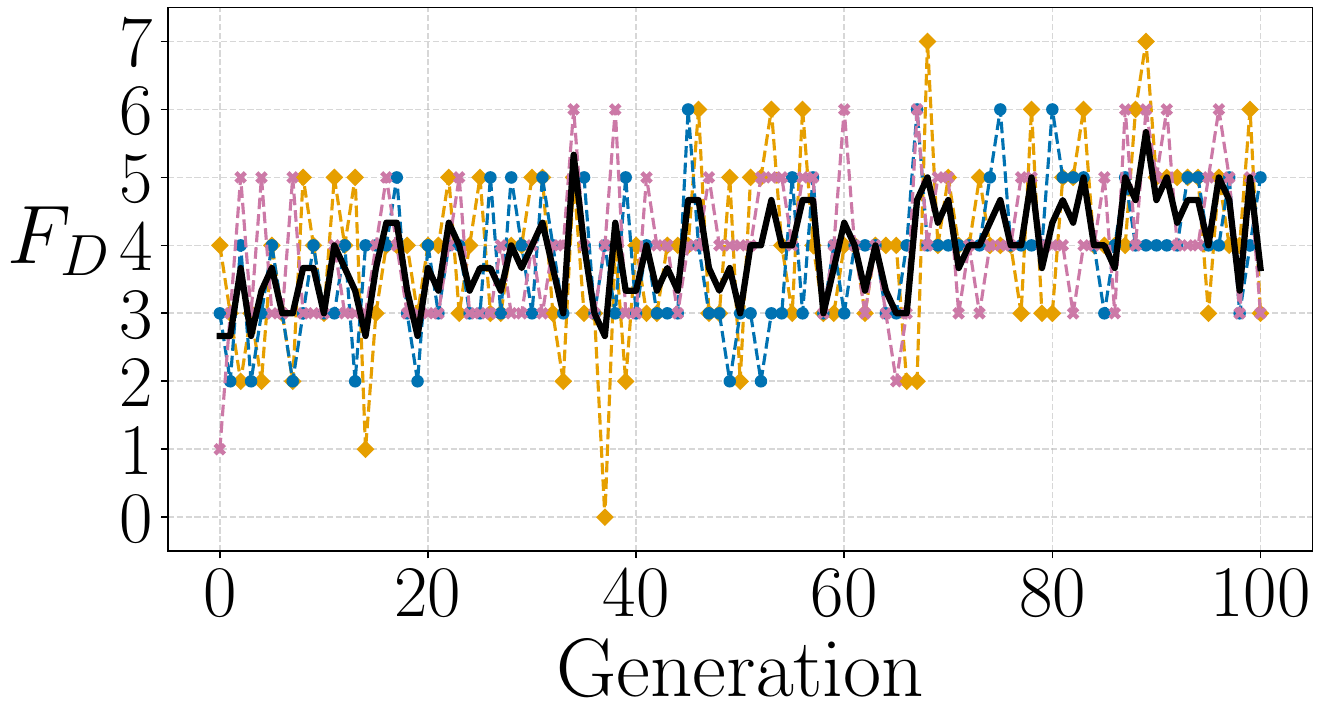}} \hfill 
\subfloat[collector \label{fig:fit_collector}]{\includegraphics[width=0.33\textwidth]{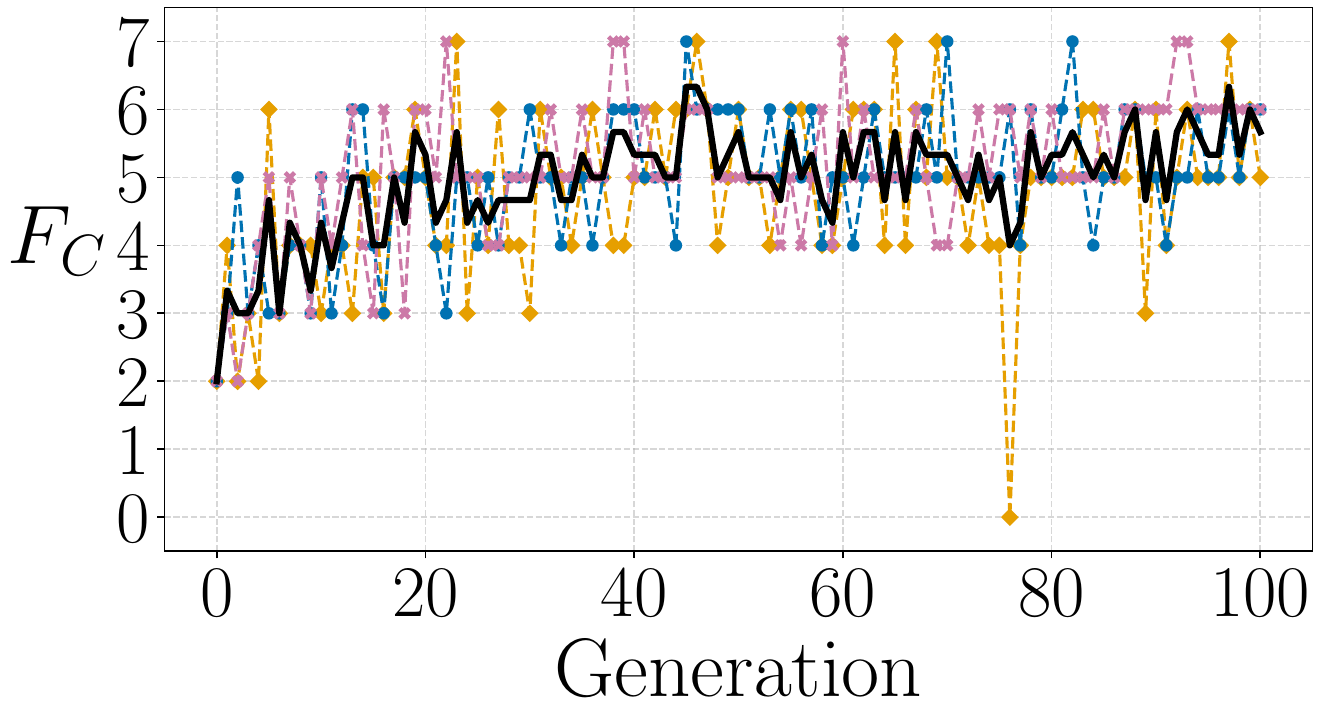}} \hfill
\caption{Best fitness (Eqs.~\ref{eq:F_G},~\ref{eq:F_D}) for the generalists, droppers, and collectors over 100~generations. Dashed lines: different independent evolutionary runs; black line: mean. 
}
\label{fig:evolution_controllers}
\end{figure}

All three behaviors (see Sec.~\ref{sec:task_description}) are successfully evolved. 
Generalists require \(30\)~generations before consistently achieving non-zero fitness, reaching a mean fitness of~\(5\) for the best evolved individuals of the last generation.
Within \(30\)~generations, droppers and collectors steadily reach mean scores of~\(4\) and~\(4.7\), respectively.
However, their fitness curve is flatter than for the generalist (see Fig.~\ref{fig:evolution_controllers}). 
These differences likely stem from the initial conditions and behavior complexity.
Generalists are initialized in the nest and must first learn to navigate to the source.
Since our aggregate fitness function~\(F_G\) does not reward intermediate progress toward the source, evolution takes several generations before object collection reliably emerges.
By contrast, droppers and collectors are initialized close to the objects, so even circular movements can yield non-zero fitness early on. This may, however, allow less robust behaviors to persist.

\subsection{Post-Evaluation}
\label{subsec:posteval_results}

\begin{figure}[tb]
\centering
\includegraphics[width=0.45\textwidth]{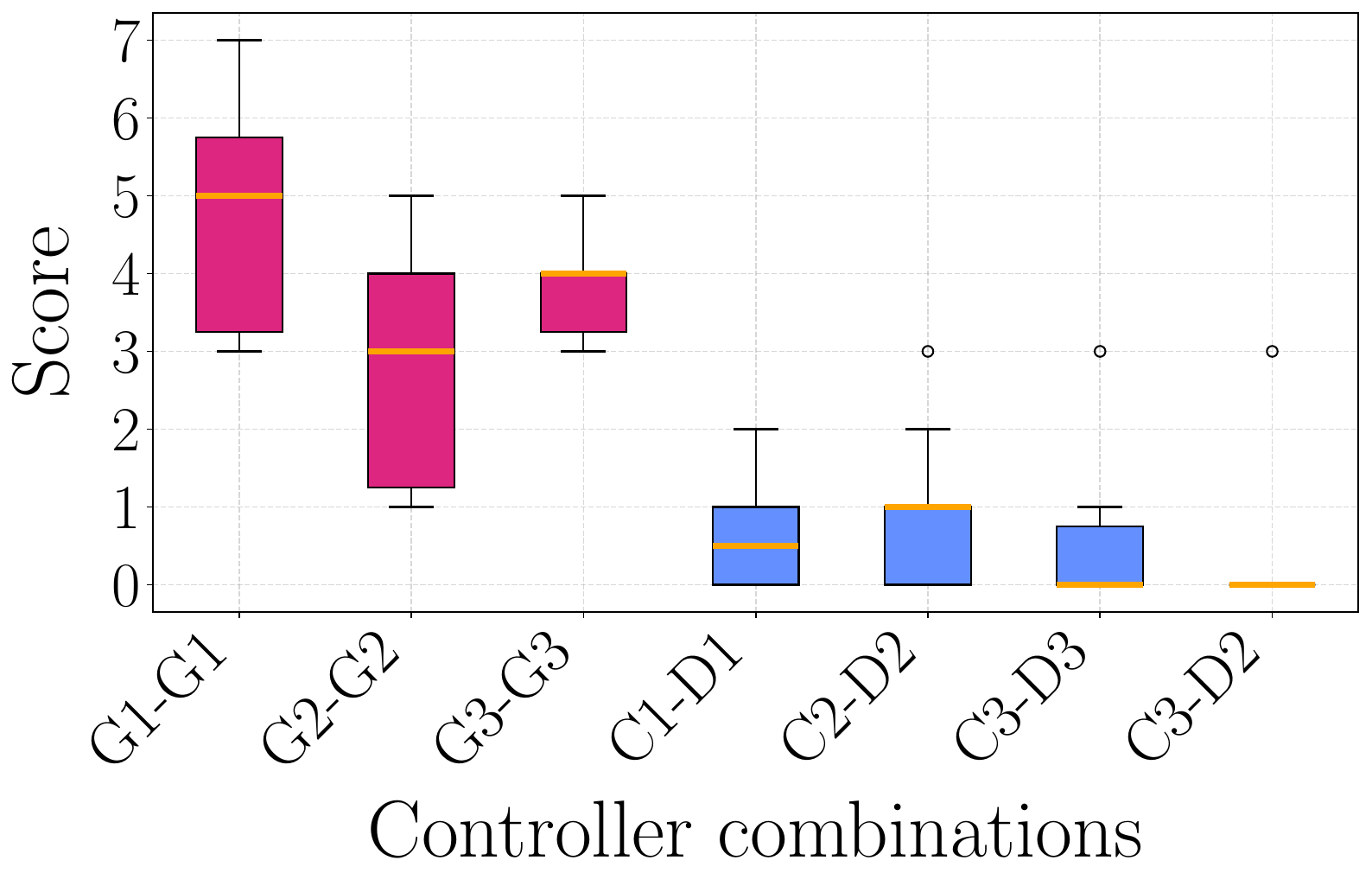}
\caption{Performance of groups of \(2\)~robots.  
Letters indicate the subtask performed (generalist~G, dropper~D, collector~C), numbers indicate the evolutionary run. 
Each \mbox{letter-number} pair is associated with a robot. 
} 
\label{fig:multirobot_score}
\end{figure}

Next, we post-evaluate the three evolved behaviors in a multi-robot scenario (see Sec.~\ref{sec:post-evaluation}).
Letters indicate the behavior with generalists~G, droppers~D, collectors~C, and numbers the evolutionary run of the respective best individual. 
The overall best evolved dropper emerged in evolution~\(2\). 
The collectors~C2 and C3 have equal performance.
For the evaluation of the overall best evolved individuals, we consider the combinations of D2 and C2 and of D2 and C3.

Generalists outperform task-partitioned groups (see Fig.~\ref{fig:multirobot_score}). 
The combination of the best evolved individuals, C3-D2, achieved the worst score, while C2-D2 performed the best among task-specialist groups.
The differences can be explained by the behaviors of C2 and C3. 
C2 covers the cache area more homogeneously, while C3 shows circular movements at one side of the cache. 
Consequently, C2 may collect objects that spread over the entire cache, making it more reliable than C3. 
The performance gap between generalist and task\hyp{}specialist groups mainly stems from the interdependence between the specialist subtasks.
Both dropper and collector must perform reliably, while poor performance in either directly limits maximum achievable system performance.
For example, even if the dropper delivers all objects quickly to the cache, success still depends on the collector bringing them to the nest.
In contrast, generalists can complete the full task independently and contribute directly to system performance. 
Moreover, generalists must find and grasp each object once, whereas specialists require both, dropper and collector, to do so. 
Limiting these actions can improve performance if the dropper-collector cooperation is inefficient.

Another factor is the difference between the evolutionary setup, which employs one robot, and the multi-robot scenario, where two robots share the same arena. 
For generalists, this can reduce performance as robots may interfere with each other.
Droppers and collectors operate at opposite ends of the same arena. 
If a robot leaves its area, it encounters sensor inputs not seen during optimization, which leads to unexpected behaviors and eventually reduces system performance.

\section{Discussion and Conclusion}

Task specialization can improve efficiency in multi-robot systems.
However, our results show that its optimization cost does not always outweigh the benefits of the task decomposition.
Under a fixed evaluation budget, generalist behaviors can outperform task-specialized controllers due to 
robot-environment and robot-robot interactions.
Alternative evolutionary settings or more advanced controllers~\cite{hiraga2018evolving}
may improve behavior quality, but better overall performance is not guaranteed.
In our scenario, object finding and grasping seems to be challenging and is required twice as often in the task-specialized case, which can reduce system performance. 
This supports Nolfi's claim~\cite{nolfi98} that monolithic approaches may outperform decomposition and integration approaches when the overall behavior emerges from complex interactions.

In future work, we will study whether larger evaluation budgets enable more effective task-specialized behaviors. 
We aim to conduct a comprehensive cost\hyp{}benefit analysis comparing generalist and task-specialized behaviors. 
To accelerate evolution, we will switch to a more scalable simulator~\cite{calderon2022swarm}, and address the potentially larger sim\hyp{}to\hyp{}real gap via a multi-level modeling approach~\cite{baumann22}, progressing from low to high-fidelity simulators and finally to real robots.

\subsubsection{Acknowledgements}
PL, HH, and JK acknowledge support from DFG through Germany's Excellence Strategy-EXC 2117-422037984 and Centre for the Advanced Study of Collective Behaviour (CASCB), University of Konstanz, Konstanz, Germany.
JK acknowledges support from the Zukunftskolleg and the Carl-Zeiss-Foundation.

\subsubsection{Disclosure of Interests}
The authors have no competing interests to declare that are relevant to the content of this article. 
%
%
%
\bibliographystyle{splncs04}
\bibliography{additions}
%

%
\end{document}